\documentclass[journal]{IEEEtran}
\usepackage{breqn}
\usepackage{amsmath}
\usepackage{algorithm}
\usepackage{amssymb}
\usepackage{algorithmic}
\usepackage{multirow}

\ifCLASSINFOpdf
\else
\fi
\begin{document}
%
\title{GAN-Based Object Removal in High-Resolution Satellite Images}
%
%
%

\author{Hadi~Mansourifar
        and~ Steven J.~Simske
\thanks{Steven J. Simske is with the Department of Systems Engineering, Colorado State University, Fort Collins, CO 80523, USA }  
}

\maketitle

\begin{abstract}

 Satellite images often contain a significant level of sensitive data compared to ground-view images. That is why satellite images are more likely to be intentionally manipulated to hide specific objects and structures. GAN-based approaches have been employed to create forged images with two major problems: (i) adding a new object to the scene to hide a specific object or region may create unrealistic merging with surrounding areas; and (ii) using masks on color feature images has proven to be unsuccessful in GAN-based object removal. In this paper, we tackle the problem of object removal in high-resolution satellite images given a limited number of training data. Furthermore, we take advantage of conditional GANs (CGANs) to collect perhaps the first GAN-based forged satellite image data set. All forged instances were manipulated via CGANs trained by Canny Feature Images for object removal. As part of our experiments, we demonstrate that distinguishing the collected forged images from authentic (original) images is highly challenging for fake image detector models. 
 
\end{abstract}


%
\IEEEpeerreviewmaketitle

\section{Introduction}

\IEEEPARstart{I}{nteractive} editing of ground-view images to remove objects has been practiced before. Object removal is a significantly important task with a variety of applications such as data augmentation \cite{mansourifar2019virtual} and content filtering for privacy and security issues \cite{kasichainula2021poisoning}. However, there are fundamental differences between satellite images and ground-view pictures \cite{everingham2015pascal}. In satellite imagery, every pixel has semantic meaning. However, ground-view images include a few foreground objects of interest with an essentially meaningless background \cite{audebert2016semantic}. This is the core problem in satellite imaging, which is why semantic segmentation and object detection are challenging. Furthermore, satellite images can cover a wide range of region types with sensitive objects and structures having a wide range of digital attributes (sharpness, edge field characteristics, entropy, etc.). This characteristic makes the satellite images potential targets for image forgery.
GAN-based satellite imaging \cite{mansourifar2022gan} has been widely used and led to
considerable progress in various imaging tasks over the past few years. Accordingly, GAN-based object removal has been practiced before. While this progress is remarkable, training a GAN needs a huge amount of training data, which makes it impossible to be used for high-quality object removal in a single high-resolution satellite image. The key challenges in this context are as follows:
\begin{itemize}
\item Few-shot GAN training for object removal.
\item Seemless object removal.
\item Low response time for real-time GAN training and generating the output.
\item The minimum degradation of reproduced regions in the scene.
\end{itemize}

To the best of our knowledge, such a scenario has never been investigated before, and we take the first step in this direction:
We slice a high-resolution satellite image into a set of 256 * 256 images.
We train a Pix2pixHD model as a Conditional GAN (CGAN) using the Canny Feature Image (CFI) of training data.
We demonstrate that CFI can outperform Segmented Feature Images (SFI) in CGAN-based object removal.
We use object detection models to evaluate the quality of generated forged images. This step is required to evaluate the object removal impact on overall image degradation. Any distortion in the image after object removal can impact the ability of object detection models to detect the rest of the target objects.
We test several classifiers to test if our GAN-generated forged images can be detected by binary classifiers.
To do so, we train several classification methods, including binary classifier and transfer learning.
\begin{figure}[]
\centering
  \includegraphics[width=50mm]{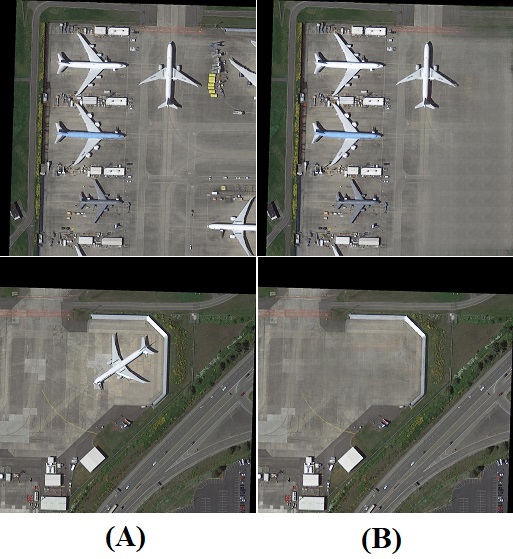}
  \caption{Instances of object inclusion (A) and object removal outputs (B) using the proposed method. }
  \label{1}
\end{figure}

 Our contributions are as follows.

\begin{itemize}
\item We propose one-shot object removal for forged satellite image generation using CFI.
\item We demonstrate the capability of CFI in the absence of rare SFI for GAN-based object removal.

\item We publish a new forged image dataset to be used in future forged image detection research. 
\item We run extensive experiments to compare the impact of CFI and SFI in image degradation after object removal.
\end{itemize}

The rest of this paper is organized as follows. Section II presents the related works. Section III introduces the proposed method. Section IV demonstrates the experiments. Section V presents the evaluation results. Finally, section VI concludes the paper.

\section{Related Works}
GAN-based forged satellite image generation and detection is a new research area in deep learning and training, with only limited works found in the literature. Yarlagadda et al. \cite{yarlagadda2018satellite} took the first step in this direction. They proposed an algorithm for satellite image forgery detection and localization under the assumption that no forged images are available for training. To do so, they used a one-class support vector machine (SVM) as an anomaly detector and trained this on feature representations of pristine satellite images obtained by GAN.
Bartusiak et al. \cite{bartusiak2019adversarial} proposed the detection and localization
of splicing in satellite images which refers to the replacement
of pixels of a region of the image to add or remove an
object. They employed a Conditional Generative Adversarial
Network (CGAN) to learn a mapping from a satellite image
to its splicing mask. The trained CGAN operates on a
satellite image of interest and outputs a mask of the same
resolution that indicates the likelihood of a pixel belonging
to a spliced region. Their proposed CGAN architecture extends the popular pix2pix \cite{isola2017image}. This approach is different
from \cite{yarlagadda2018satellite} since they used both pristine and spliced images (i.e., forged images not employed by \cite{yarlagadda2018satellite}) to train the model. Horvath et al. \cite{horvath2019anomaly} proposed a deep learning based method for detecting and localizing splicing manipulations.
In \cite{horvath2020manipulation}, a one-class detection
method was proposed based on deep belief networks (DBN) for splicing
detection and localization without using any prior knowledge of the manipulations. Montserrat et al. \cite{montserrat2020generative} used
two generative autoregressive models, PixelCNN \cite{van2016pixel}, and Gated PixelCNN \cite{van2016conditional}, to detect pixel-level manipulations. These neural networks were originally devised to generate new images by modeling a pixel's distribution given a set of neighboring pixels and a conditional likelihood value assigned to each pixel. Similarly, the pixels with a low likelihood scoring by the neural network can be detected as manipulated ones. \cite{horvath2021manipulation} proposed an unsupervised technique that uses a
so-called "Vision Transformer" to detect spliced areas within satellite images.

\section{Background}
\subsection{Vanilla GAN}
The initial version of GANs \cite{goodfellow2014generative} is known as Vanilla GAN. The learning process of the Vanilla GANs is to train a discriminator $D$ and a generator $G$ simultaneously. The target of $G$ is to learn the distribution $p_g$ over data $x$. $G$ starts from sampling input variables $z$ from a uniform or Gaussian distribution $p_z(z)$, then maps the input variables $z$ to data space $G(z; \theta_g)$ through a differentiable network where $\theta_g$ represents network parameters. $D$ is a classifier $D(x; \theta_d)$ that aims to recognize whether the input is from training data or from $G$ where $\theta_d$ represents network parameters. The minimax objective for GANs can be formulated as follows:
\begin{dmath}
_{G}^{min} \quad_{D}^{max} \quad V_{GAN} (D,G) = \mathbb{E} _ {x \sim p_{x} } [log  D(x)]  + \quad \mathbb{E} _ {z \sim q_{z} } [log  (1- D(G(z)))] 
\end{dmath}
\subsection{Conditional GAN}
In Conditional GAN (CGAN) \cite{mirza2014conditional}, labels act as an extension to the latent space $z$ to generate and discriminate images better.
The objective function of CGANs is as follows:
\begin{dmath}
\min_G\max_DV(D,G)=\mathbb{E}_{x\sim p(data)(x)}[\log D_({_X|_Y} _)]+
\mathbb{E}_{Z\sim pz}(z))[\log (1-D(G(_Z|_Y)))]
\end{dmath}

\subsection{Pix2Pix}
Pix2pix \cite{isola2017image} image is a conditional GAN that uses feature images as labels to be translated to target images. Pix2pix uses U-Net \cite{zhang2018road,kohl2018probabilistic} as the architecture of generator PatchGAN for discriminator architecture. The most important feature of the Pix2pix generator is skip connections between each layer $i$ and layer $n - i$ to concatenate all channels at layer $i$ with those
at layer $n - i$, where $n$ is the total number of layers.

\subsection{Pix2PixHD}
To improve the pix2pix framework, \cite{wang2018high} used a coarse-to-fine generator, a multi-scale discriminator architecture, and
a robust adversarial learning objective function. The generator in Pix2pixHD consists of two sub-networks, $G_1$, and $G_2$, where $G_1$ is the
global generator network, and $G_2$ is the local enhancer network. As a multi-scale discriminator architecture, 3 discriminators with an identical network structure operate at different image scales. The real and synthesized high-resolution images are down-sampled by a factor of 2 and 4 to create an image pyramid of 3 scales. The discriminators $D_1$, $D_2$, and $D_3$ are
then trained to differentiate real and synthesized images at the 3 different scales, respectively.

\begin{figure*}[]
\centering
  \includegraphics[width=145mm]{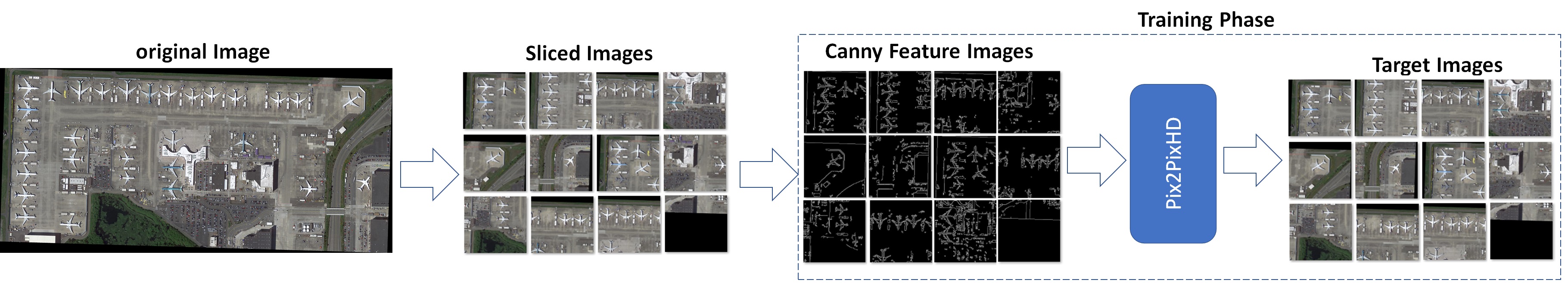}
  \caption{Training pipeline of CGAN-based object removal . }
  \label{3}
\end{figure*}

\begin{figure*}[]
\centering
  \includegraphics[width=145mm]{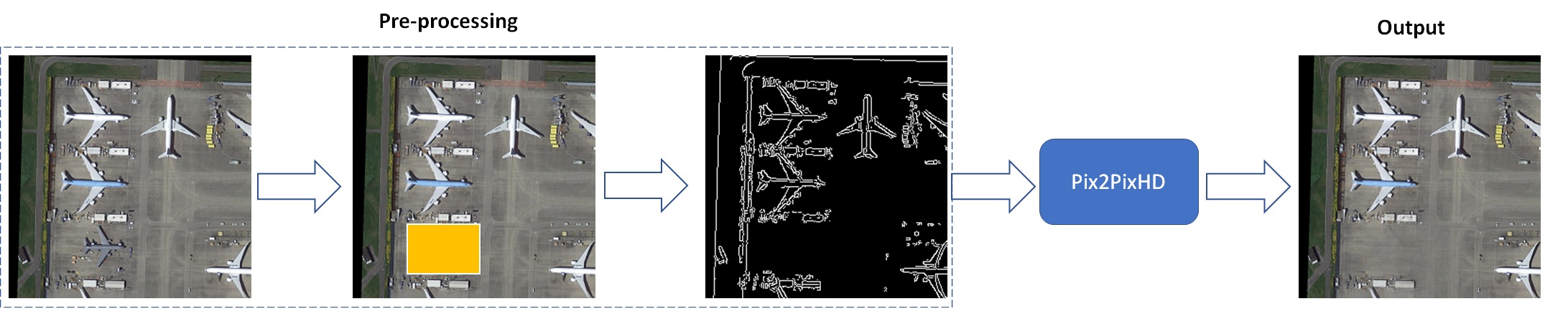}
  \caption{Inference pipeline of CGAN-based Object Removal.}
  \label{2}
\end{figure*}
\section{Proposed Method}
In this section, we demonstrate our object removal approach in high-resolution satellite images. The proposed method is divided into training and inference phases which are demonstrated in the following sections.
\subsection{Training phase}
The goal of this phase is to train a CGAN like traditional Pix2pix or Pix2pixHD given one single high-resolution satellite image, as shown in Figure \ref{3}. The following Algorithm shows the required steps.

\begin{algorithm}[H]
\caption{The training phase of proposed CGAN-based object removal}
\begin{algorithmic}[1]
       \STATE{Slice the input high-resolution satellite image into a set of 256*256 blocks.}
      \STATE{Extract the CFI of each sliced block of Step 1.} 
      \STATE{ Train a Pix2pixHD model to translate CFI to each corresponding training image. } 
      \STATE{Train pix2pixHD to translate obtained images from Step 2 to output image.}

\end{algorithmic}
\end{algorithm}

\subsection{Inference phase}
In this phase, the user can define a region on the original image then the edges represented by white are removed in the Canny feature image. Finally, the edited feature image is passed to the trained Pix2pixHD, as shown in Figure \ref{2}. The following Algorithm shows the required steps.

\begin{algorithm}[H]
\caption{The inference phase of proposed CGAN-based object removal }
\begin{algorithmic}[1]
       \STATE{Extract the Canny edges of a target block for object removal as CFI.}
      \STATE{Define the target region for object removal.} 
      \STATE{Remove white pixels in the defined region of Step 2 from CFI.}
      \STATE{Translate obtained image from Step 3 to the output image using the trained pix2pixHD.}

\end{algorithmic}
\end{algorithm}

\subsection{Advantages of One-Shot Object Removal}
Two major problems of training GANs in the absence of sufficient training data are the lack of generalization and the lack of diversity. The former is caused by
over-fitting, and the latter is due to mode collapse. \cite{shaham2019singan,hinz2021improved}. While over-fitting is problematic when embedding a new object in the image, over-fitting can play a positive role in removing weak objects from the scene. From the semantic segmentation point of view, most pixels are categorized as background. By default, any deep neural network for image segmentation or image synthesis has a bias toward the background pixels. This makes weak object augmentation a particular challenge. However, weak object removal is made easier by the CGAN, since the trained CGAN is iteratively rewarded for replacing the weak object pixels with the background pixels. A CGAN trained by one single instance has even more capability for object removal since it has more bias toward most pixels. To do so, we start with a single satellite image, slice it to 256 * 256 images and extract the corresponding Canny feature images to train a Pix2pixHD GAN as shown in Figure \ref{3}.

\subsection{The Advantage of Canny Feature Images}
While polygon-level masks are widely used for object detection tasks, pixel-level masks are a more proper choice for object augmentation-removal applications. In cases of object removal, we need feature images comprising minimal information about the objects and structures so that they can exacerbate the over-fitting, which is required for high-quality object removal given only a single image. 

\begin{figure*}[]
\centering
  \includegraphics[width=120mm]{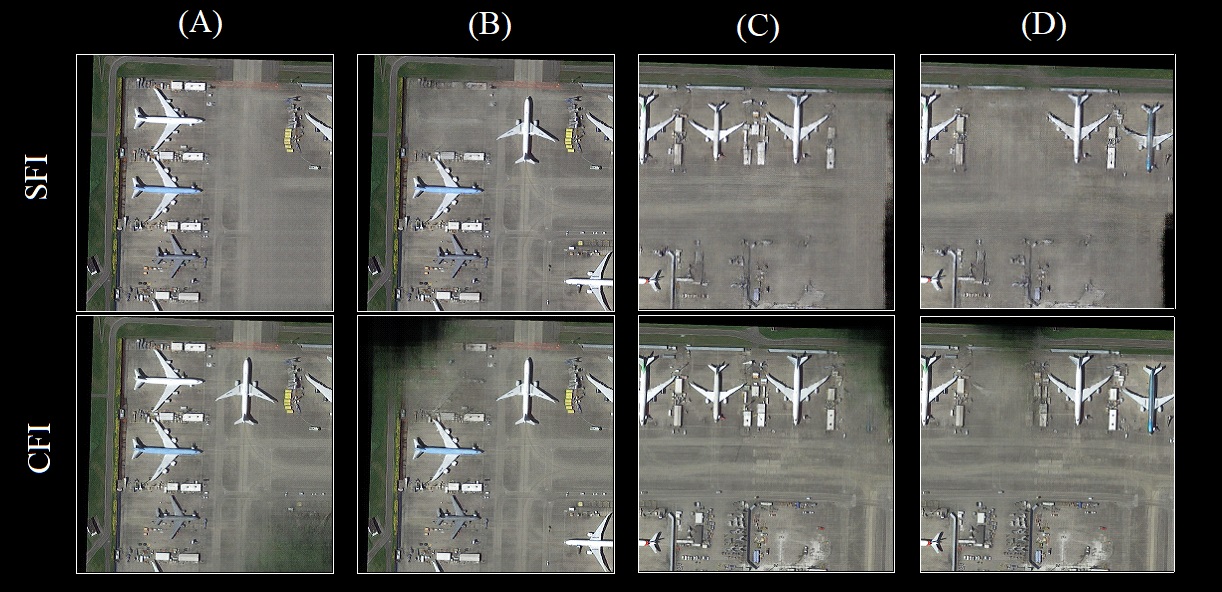}
  \caption{CGAN-based object removal: Canny feature images versus segmented feature images.  }
  \label{4}
\end{figure*}

\section{Experiments}
In this section, we present the experiment setup, architectures, dataset, metrics, and results corresponding to forged image generation and detection.
\subsection{Metrics}
We used two sets of metrics as follows.
\begin{itemize}
\item Similarity Scores: including MSE, PSNR, and SSIM to evaluate the degradation 
\item Object Detection Scores: to compare the CFI and SFI impact on object detection scores.
\end{itemize}
\subsection{CFI Versus SFI for CGAN-based Object Removal}
CGAN-based image manipulation suffers a significant problem: any change in a local region of a feature image can impact the other blocks in the final output. That's why we are interested in evaluating the local and global impact of CFI and SFI manipulation. The best way is to use image similarity metrics to compare the ground truth and output image. The goal of these experiments is to test the capability of CFI to be used in CGAN-based object removal in the absence of pixel-wise annotated SFI, which is rare in the satellite imaging domain. Although the ground truth and output image are not exactly the same due to object removal, similarity scores can still help to compare CFI and SFI. Table I shows the comparison results of output images shown in Figure 4 and the corresponding original ground truths. Table II also tabulates the object detection scores.
\begin{table}[H]
\centering
\caption{Comparison of CFI and SFI using image similarity scores}
\begin{tabular}{l|l|l|l|l|l|}
\cline{2-6}
 &  & \textit{\textbf{A}} & \textit{\textbf{B}} & \textit{\textbf{C}} & \textit{\textbf{D}} \\ \hline
\multicolumn{1}{|l|}{\multirow{2}{*}{\textbf{MSE}}} & \textit{\textbf{CFI}} & \textbf{0.129} & \textbf{0.085} & 0.248 & 0.253 \\ \cline{2-6} 
\multicolumn{1}{|l|}{} & \textit{\textbf{SFI}} & 0.137 & 0.163 & \textbf{0.132} & \textbf{0.135} \\ \hline
\multicolumn{1}{|l|}{\multirow{2}{*}{\textbf{PSNR}}} & \textit{\textbf{CFI}} & \textbf{32.95} & \textbf{34.73} & 30.10 & 30.02 \\ \cline{2-6} 
\multicolumn{1}{|l|}{} & \textit{\textbf{SFI}} & 32.69 & 31.93 & \textbf{32.84} & \textbf{32.7} \\ \hline
\multicolumn{1}{|l|}{\multirow{2}{*}{\textbf{SSIM}}} & \textit{\textbf{CFI}} & 0.844 & \textbf{0.92} & 0.69 & 0.66 \\ \cline{2-6} 
\multicolumn{1}{|l|}{} & \textit{\textbf{SFI}} & \textbf{0.901} & 0.83 & \textbf{0.887} & \textbf{0.87} \\ \hline
\end{tabular}
\end{table}

\begin{table}[H]
\centering
\caption{Object detection scores returned by AWS Rekognition.}
\begin{tabular}{|c|c|c|c|c|c|c|c|c|} 
\cline{2-9}
\multicolumn{1}{c|}{} & \multicolumn{4}{c|}{\textit{\textbf{Airplane}}}   & \multicolumn{4}{c|}{\textit{\textbf{Aircraft}}}    \\ 
\cline{2-9}
\multicolumn{1}{c|}{} & \textbf{A} & \textbf{B} & \textbf{C} & \textbf{D} & \textbf{A} & \textbf{B} & \textbf{C} & \textbf{D}  \\ 
\hline
CFI                   & 99.3       & 99.2       & 97.5       & 87.8       & 99.3       & 99.2       & 97.5       & 87.8        \\ 
\hline
SFI                   & 99.3       & 99.2       & 97.4       & 97.5       & 99.3       & 99.2       & 97.4       & 97.5        \\ 
\hline
\multicolumn{1}{c|}{} & \multicolumn{4}{c|}{\textit{\textbf{Terminal}}}   & \multicolumn{4}{c|}{\textit{\textbf{Vehicle}}}     \\ 
\cline{2-9}
\multicolumn{1}{c|}{} & \textbf{A} & \textbf{B} & \textbf{C} & \textbf{D} & \textbf{A} & \textbf{B} & \textbf{C} & \textbf{D}  \\ 
\hline
\textit{CFI}          & 0          & 55.9       & 55.1       & 0          & 99.3       & 99.2       & 97.5       & 87.8        \\ 
\hline
\textit{SFI}          & 55.7       & 55.6       & 69.2       & 56.3       & 99.3       & 99.2       & 97.4       & 97.5        \\
\hline
\end{tabular}
\end{table}

The results shown in Table I and Table II demonstrate that CFI is robust enough for CGAN-based  object removal since we can observe the superiority of CFI over SFI in some cases. Besides, in the majority of failed cases, the CFI results are competable with SFI results.
\subsection{Dataset}
In order to train satellite forged image detection models, we collected a dataset by CGAN-based object removal using CFI. To do so, we manipulated instances of the iSAID dataset \cite{Xia_2018_CVPR,waqas2019isaid}. Existing satellite image datasets are either well-suited for semantic segmentation or object detection. iSAID is the first benchmark dataset, for instance, segmentation in aerial images. This large-scale and densely annotated dataset contains 655,451 object instances for 15 categories across 2,806 high-resolution images. The collected dataset contains 162 forged images and 266 pristine images as training data. The validation set includes 95 forged images and 114 pristine images.
\begin{figure*}[]
\centering
  \includegraphics[width=135mm]{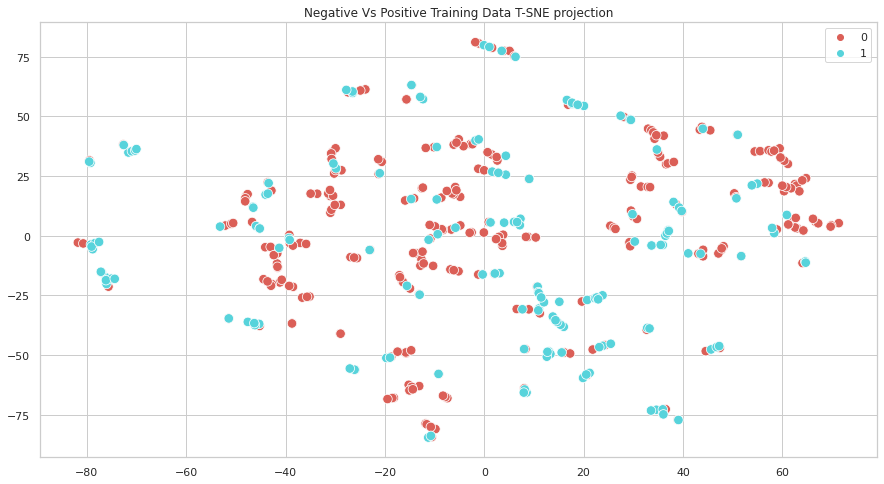}
  \caption{TSNE data projection: positive versus negative data distribution.}
  \label{8}
\end{figure*}
\subsection{Forged Satellite Image Detection}
\begin{figure}[]
\centering
  \includegraphics[width=60mm]{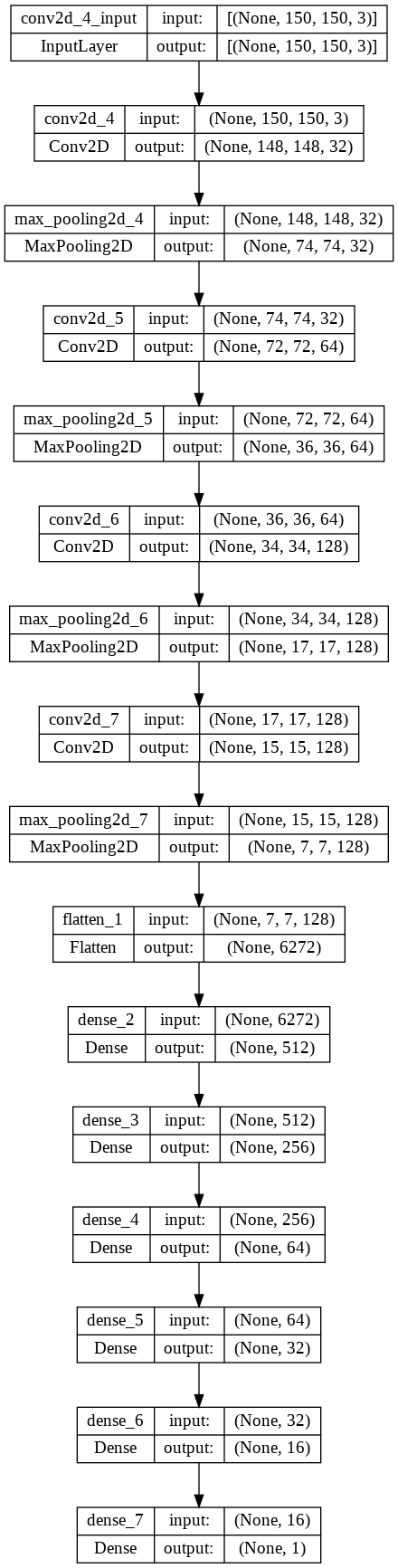}
  \caption{The architecture of binary forged image detector. }
  \label{9}
\end{figure}

Detecting the regions manipulated by the proposed object removal method is not trivial using traditional anomaly detection methods for two reasons: (i) The range of reconstruction losses or anomaly scores are so different in manipulated, and original images due to the diversity of satellite view scenes; and (ii) In many manipulated images, the anomaly score is lower than for the original images, since removing the objects may lead to lower entropy. Due to the aforementioned conditions, defining a threshold to detect manipulated images with removed objects takes a lot of work. In this section, we test two different methods, including binary classification and transfer learning to detect forged satellite images. The goal of this experiment is to test whether the forged images are distinguishable by the binary classifiers easily.
\begin{itemize}
\item Binary Forged Image Detector: It's a convolution neural network with ten layers (as shown in Figure \ref{9}), Adam optimizer with learning rate=0.001 and 100 epochs for training. 

\item Transfer Learning: In this approach, MobileNetV2 \cite{sandler2018mobilenetv2} is fine-tuned with the training data with 100 epochs and RMSprop as optimizer.
\end{itemize}
The ROC curves of both models show their failure to distinguish the collected forged images from pristine instances as shown in Figure \ref{10}. 
\subsection{Dataset Visualization}
\begin{figure}[]
\centering
  \includegraphics[width=65mm]{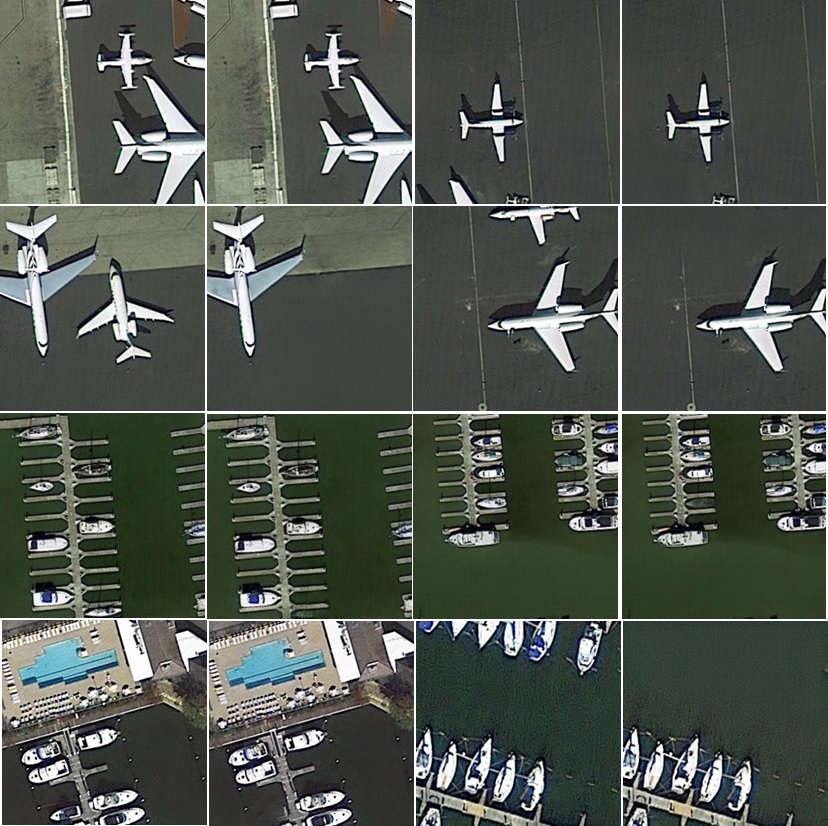}
  \caption{Some forged samples of collected dataset. }
  \label{7}
\end{figure}

\begin{figure}[]
\centering
  \includegraphics[width=75mm]{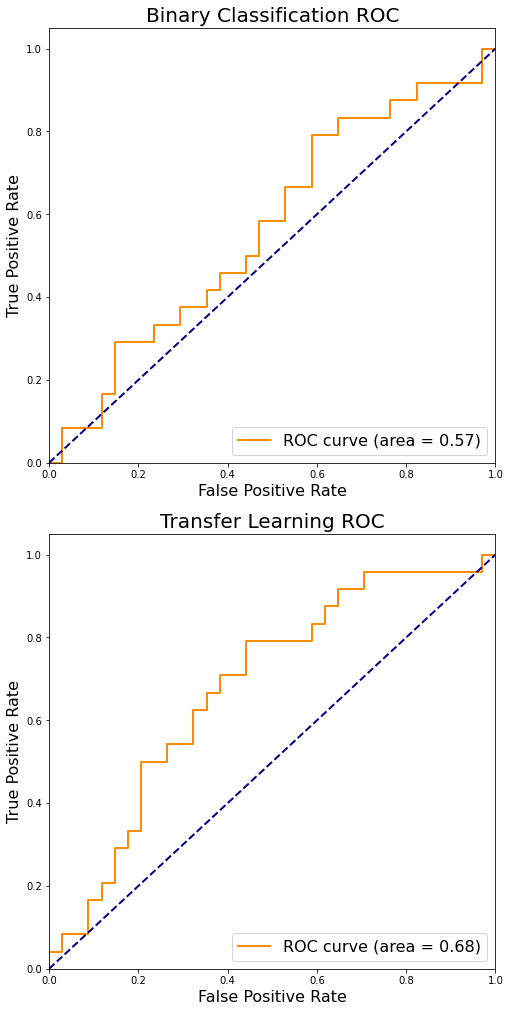}
  \caption{Receiver operating characteristic curve corresponding to the binary classifier and transfer learning for forged image detection.  }
  \label{10}
\end{figure}

Figure \ref{7} shows some of the collected forged instances using CGAN-based object removal. Furthermore, Figure \ref{8} shows the TSNE projection of positive-negative collected instances. To calculate the TSNE data projection, first, we passed the images to ResNet50 \cite{mukti2019transfer} to reduce the dimensionality by collecting 1000 label probabilities.

\section{Conclusion}
In this paper, we investigated the capabilities of CFI for one-shot CGAN-based object removal. Full or even partial pixel-wise annotation is rare in existing satellite imaging datasets. In the absence of pixel-wise annotation, Canny Feature Image (CFI) is an inevitable choice for CGAN-based object removal. We successfully compared the CFI and SFI to show that CFI is robust enough to be used in CGAN-based image manipulation.

\bibliographystyle{IEEEtran}
\bibliography{ref.bib}
\end{document}